\documentclass[letterpaper, 10pt, conference]{IEEEtran}

\usepackage{fancyhdr}
\fancypagestyle{firstpage}{
  \fancyhf{} 
   
  \fancyhead[l]{\small \textit{Accepted at The 22nd IEEE International Conference on Mobile Ad-Hoc and Smart Systems 2025\\ Camera‑ready version will submit on August 15, 2025}} 
}

\IEEEoverridecommandlockouts

\usepackage{cite}
\usepackage{amsmath,amssymb,amsfonts}
\usepackage{algorithmic}
\usepackage{graphicx}
\usepackage{textcomp}
\usepackage{multirow}
\usepackage[utf8]{inputenc}
\usepackage[bottom]{footmisc}
\usepackage{amsmath}
\usepackage{caption}
\usepackage{subcaption}
\usepackage{mathtools} 
\usepackage{tablefootnote}
\usepackage[euler]{textgreek}
\usepackage{tikz}
\usepackage{booktabs} 
\usepackage[utf8]{inputenc}
\usepackage[most]{tcolorbox}
\usepackage{enumitem}
\definecolor{systemframe}{HTML}{558B2F}  
\definecolor{systemback}{HTML}{C5E1A5}  
\definecolor{userframe}{HTML}{558B2F}  
\definecolor{userback}{HTML}{C5E1A5}  
\usepackage{adjustbox}
\usepackage{graphicx}
\usepackage[bookmarks=false, pdfencoding=auto, pdfborder={0 0 0}, pdfusetitle, draft]{hyperref}

\usepackage{pifont}
\usepackage{circledsteps}
\tcbset{
  jsonbox/.style={
    colback=gray!10,
    colframe=gray!40,
    boxrule=0.4pt,
    arc=1mm,
    fontupper=\ttfamily\footnotesize,
    sharp corners,
    left=2mm, right=2mm, top=1mm, bottom=1mm
  }
}
\usepackage{listings}
\def\BibTeX{{\rm B\kern-.05em{\sc i\kern-.025em b}\kern-.08em
    T\kern-.1667em\lower.7ex\hbox{E}\kern-.125emX}}
\begin{document}

\title{DailyLLM: Context-Aware Activity Log Generation Using Multi-Modal Sensors and LLMs}

\author{}

\author{\IEEEauthorblockN{Ye Tian, Xiaoyuan Ren, Zihao Wang, Onat Gungor, Xiaofan Yu and Tajana Rosing }
\IEEEauthorblockA{\textit{University of California San Diego, Computer Science and Engineering Department} \\
\{yet002, x6ren, ziw140, ogungor, x1yu, tajana\}@ucsd.edu}
}

\maketitle
\thispagestyle{firstpage}

\newcommand{\systemname}{{DailyLLM}~}
\newcommand{\systemnamenospace}{{DailyLLM}}

\begin{abstract} 

Rich and context-aware activity logs facilitate user behavior analysis and health monitoring, making them a key research focus in ubiquitous computing. The remarkable semantic understanding and generation capabilities of Large Language Models (LLMs) have recently created new opportunities for activity log generation. However, existing methods continue to exhibit notable limitations in terms of accuracy, efficiency, and semantic richness. 
To address these challenges, we propose {\systemnamenospace}. To the best of our knowledge, this is the first log generation and summarization system that comprehensively integrates contextual activity information across four dimensions: location, motion, environment, and physiology, using only sensors commonly available on smartphones and smartwatches.
To achieve this, {\systemnamenospace} introduces a lightweight LLM-based framework that integrates structured prompting with efficient feature extraction to enable high-level activity understanding. Extensive experiments demonstrate that {\systemnamenospace} outperforms state-of-the-art (SOTA) log generation methods and can be efficiently deployed on personal computers and Raspberry Pi. Utilizing only a 1.5B-parameter LLM model, {\systemnamenospace} achieves a 17\% improvement in log generation BERTScore precision compared to the 70B-parameter SOTA baseline, while delivering nearly 10$\times$ faster inference speed.

\end{abstract}


\section{Introduction}

Over the past decade, the adoption of smart mobile devices has grown substantially, with the number of global users exceeding 6.7 billion for smartphones and 224 million for smartwatches by the end of 2024 \cite{statistaSmartphone}. These numbers are expected to continue growing steadily. The ubiquity of smart portable devices enables the continuous acquisition of activity data, offering significant opportunities for fine-grained monitoring and analysis of user behaviors and health conditions.
Nonetheless, activity classification alone does not suffice for a comprehensive analysis of behavior and health. To gain deeper insights, it is equally critical to account for the temporal, spatial, environmental, and physiological dynamics associated with these activities. For example, identifying ``prolonged sitting in dim light, accompanied by a lowered heart rate, reduced body temperature, and increased blood pressure'' provides far more critical insights than simply labeling the activity as ``sitting.'' This rich context information enables personalized health interventions, facilitates lifestyle analysis, and contributes to the creation of ``digital memories'' that are valuable for users \cite{aiordachioae2019life,jiang2019memento,aiorduachioae2024integrating}. Digital memories can assist users in recalling their daily routines, which is particularly beneficial for individuals with cognitive impairments, such as those affected by Alzheimer’s disease. As a result, the generation of context-aware activity logs has emerged as a critical area of research \cite{xu2024autolife,post2025contextllm}.

\begin{figure}[!t]
    \centering
    \includegraphics[width=0.9\linewidth]{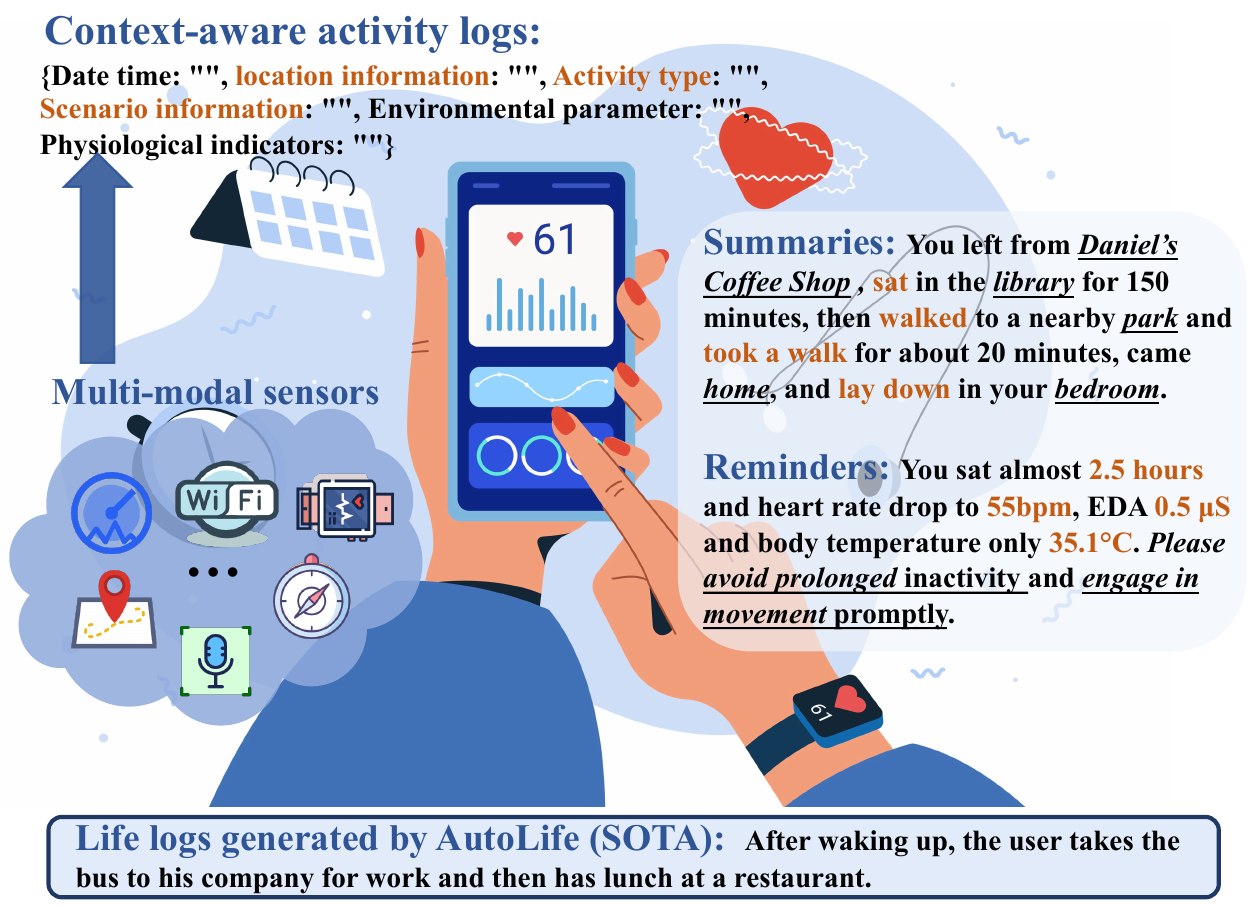}
    \vspace{-0.05in}
    \caption{DailyLLM \textit{vs.} SOTA: Improved Richness, Accuracy, and Efficiency in Activity Generation and Summarization.}
    \label{fig:first picture}
    \vspace{-0.2in}
\end{figure}


Recently, several commercial applications \cite{lifelogapp,dayOneApp} have promoted context-aware activity logging through photos and textual notes; however, they rely heavily on manual input, which imposes a considerable burden on users. Prior studies \cite{sahoo2019foodai,jeong2020cyberslacking,krieter2018analyzing,lewin2024novel,aiordachioae2019life,jiang2019memento,aiorduachioae2024integrating} have investigated smart glasses with high-resolution cameras for automated life logging, which involves the continuous capture and recording of daily activities. However, the high cost of devices and privacy concerns associated with continuous video capture pose significant barriers to their practical deployment. Motivated by these limitations, our goal is to generate context-rich activity logs using the commonly available sensors embedded in smartphones and smartwatches, such as accelerometers, GPS, and Wi-Fi. 
As illustrated in Fig. \ref{fig:first picture}, our system converts low-level sensor signals into human-readable, context-aware activity logs, incorporating details such as time, location, ambient context, activity state, and physiological indicators. Leveraging these fine-grained activity logs, the system periodically generates high-level summaries and identifies abnormal patterns to deliver personalized recommendations to users. For example, it can generate reminders like: ``You have been sedentary for over 2.5 hours, and physiological signals indicate a negative trend. Please avoid prolonged inactivity and engage in movement promptly.''

\begin{figure*}[t]
    \centering
    \includegraphics[width=0.9\linewidth]{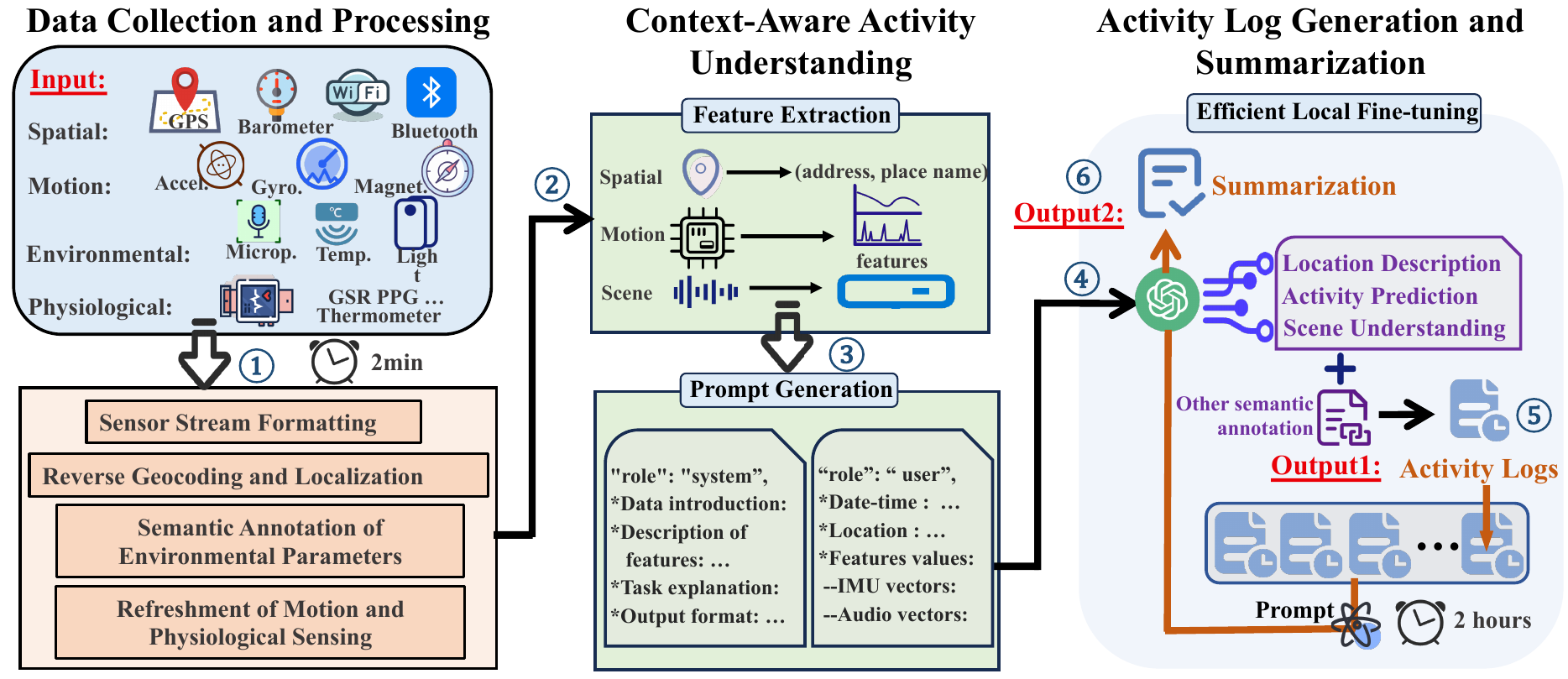}
    \caption{{\systemname}Overview, consisting of three main stages: (1) multi-modal data collection and processing, (2) context-aware activity understanding through feature extraction and prompt generation, and (3) activity log generation and summarization.}
    \label{fig:system overview}
    \vspace{-0.2in}
\end{figure*}

The semantic description and summarization of activity contexts demand a deep understanding of contextual semantics and advanced natural language generation capabilities. Large Language Models (LLMs) have exhibited significant strengths in these domains, introducing a new paradigm for activity log generation \cite{post2025contextllm}. However, despite recent efforts such as AutoLife \cite{xu2024autolife}, which demonstrated the potential of designing an LLM-based end-to-end logging system, existing approaches still exhibit limitations in terms of accuracy, efficiency, and semantic richness. The quality of the generated activity logs remains suboptimal, with the best precision reaching only 65\%, indicating substantial room for improvement.
Furthermore, system efficiency and user data privacy risks associated with cloud-based processing also present significant hurdles. To address these challenges, we propose {\systemnamenospace} (Fig. \ref{fig:system overview}), a novel LLM-driven system that leverages multi-modal sensing for the accurate and efficient generation of context-rich activity logs. It addresses the key limitations of existing approaches through a carefully designed method and system framework: 

\textbf{1) Integration and Alignment of Multi-modal Heterogeneous Sensor Data:} Generating semantically rich contextual activity information requires extracting meaningful data from a variety of heterogeneous sensors. For instance, activity location and type can be inferred from GPS and Inertial Measurement Unit (IMU) data, whereas ambient scene characteristics and physiological signals (e.g., heart rate variability) can be captured using microphones and galvanic skin response (GSR) sensors. However, these sensors differ significantly in terms of data format, sampling rate, and granularity, making effective integration and alignment a substantial challenge. To address this challenge, {\systemname}integrates a unified data processing module that synchronizes sensor readings from diverse sources and annotates them with semantic labels. This approach ensures temporal and semantic consistency, which are essential for accurate downstream inference and context reconstruction. 
\textbf{2) Generation and Summarization of High-Quality Semantic Descriptions for Contextual Activity:} Accurately transforming raw time-series sensor data into coherent, high-quality natural language descriptions presents a significant challenge.
The system must learn to translate evolving numerical patterns over extended temporal windows into human-readable event narratives and summaries. {\systemname}leverages the strong semantic capabilities of LLMs to design a context-aware activity understanding module. It designs tailored feature extraction strategies that maps low-level time-series sensor data into a semantic embedding space compatible with LLMs, enabling accurate generation of high-quality natural language descriptions.
\textbf{3) Efficient Inference Architecture and User Data Privacy Preservation:} Developing a lightweight and computationally efficient architecture is critical for enabling low-latency inference, particularly under constraint of limited resources on local devices committed to preserving user data privacy.
To address this, {\systemname}first encodes raw sensor data into compact feature representations. It then constructs structured prompts that incorporate these features along with their semantic interpretations, guiding the LLM to perform rapid, context-aware reasoning by leveraging its extensive prior knowledge. Furthermore, {\systemname}adopts Low-Rank Adaptation (LoRA) \cite{hu2022lora} to inject lightweight parameter matrices into the pre-trained LLM, enabling efficient fine-tuning with significantly reduced memory and computational overhead. These efforts facilitates deployment on personal devices, including standard PCs and resource-constrained mobile platforms such as the Raspberry Pi. 
In summary, the core contributions of this work are summarized as follows:

\begin{itemize}
    \item To the best of our knowledge, this is the first sensor-based system that can automatically generate activity logs and summaries across four key dimensions: location, motion, environment, and physiology.
    \item We construct a fine-grained and comprehensive activity context dataset, which we will publicly release to support future research in this area.
    \item Our feature extraction and prompt engineering strategies significantly improve the ability of LLMs to interpret multi-modal sensor data. {\systemnamenospace} outperforms SOTA methods, LLaSA \cite{imran2024llasa}, achieving a 12.24\% increase in F1-score for activity prediction and attaining 100\% accuracy in recognizing 15 distinct environmental scenarios.
    \item Extensive experiments demonstrate that {\systemnamenospace}, using a significantly smaller model (Deepseek-R1-1.5B), achieves a 17\% higher activity log generation BERTScore precision compared to SOTA approach with LLaMA3-70B, while delivering nearly a 10$\times$ faster inference speed.
\end{itemize}

\section{related Work}

\subsection{LLMs for Sensing Human Activity}
The potential of leveraging LLMs for more efficient reasoning tasks based on physical data has recently garnered significant attention and discussion \cite{kok2024iot,an2024iot,mo2024iot}.
Xu et al. \cite{xu2024penetrative} introduced the concept of ``Penetrative AI'' and showcased the potential of LLMs for interpreting sensor data by preliminary study. HARGPT \cite{ji2024hargpt} evaluated the ability of GPT-4 to analyze IMU data using appropriate prompting and demonstrated its capacity to perform simple HAR tasks in a zero-shot setting. Hota et al. \cite{VirtualAnnotators} further investigated the challenges GPT-4 faces in interpreting raw IMU data and utilized LLMs for annotating human activity through embedding vectors and few-shot prompting.
Recently, LLaSA \cite{imran2024llasa} fine-tuned an LLM model by integrating LIMU-BERT and Llama to interpret and respond to queries related to human activity analysis. 
However, current research on utilizing LLMs for activity recognition is still in its infancy \cite{berenguer2024using}. 
Most studies mainly focus on unimodal sensor data, classifying only a limited set of active labels, and even being able to perform simple binary classification tasks only \cite{xu2024penetrative,VirtualAnnotators}. 
In contrast, {\systemnamenospace} integrates multi-modal sensor data, capturing rich contextual information about activities, including temporal, spatial, environmental, and physiological indicators, enhancing LLM-based understanding of sensor data.

\subsection{Human Daily Life Logs Generation}
Life logging enables individuals to capture and preserve daily experiences for purposes such as self-reflection and health analysis \cite{xu2024autolife}. As a result, many early studies focused on specific scenarios, including diet tracking \cite{sahoo2019foodai}, phone usage \cite{jeong2020cyberslacking}, and sleep logging \cite{lewin2024novel}. For example, FoodAI \cite{sahoo2019foodai} estimated nutritional intake by detecting food in images, while some studies \cite{jeong2020cyberslacking} investigated mobile usage patterns and work-related behaviors through smartphone usage logs. Daniel et al. \cite{lewin2024novel} introduced an innovative voice-interactive sleep logging system that automatically provided users with sleep metrics. Recently, significant efforts have been directed towards designing life logging systems using wearable glasses. Life-Tags \cite{aiordachioae2019life} generated tag and concept clouds using smart glasses, Memento \cite{jiang2019memento} combined EEG sensors with smart glasses for emotion-driven logs, and HeadsetsLog \cite{aiorduachioae2024integrating} captured images with XR glasses and neural earphones for comprehensive life logging. However, smart glasses are expensive, and the use of cameras can raise significant privacy concerns in many scenarios. To address these issues, AutoLife \cite{xu2024autolife} relied on low-cost sensors to infer users' activity contexts and leveraged multiple existing LLMs and VLMs for life log generation. In contrast, {\systemnamenospace} integrates a broader range of multi-modal sensor data, enabling a deeper understanding of activity contexts. Moreover, {\systemnamenospace} significantly streamlines the workflow and fully deployed on personal devices, eliminating the risk of privacy user data leakage. 
\section{DailyLLM: System design}
{\systemname}designs three core modules to automatically generate activity logs and summaries from sensor data streams. As illustrated in Fig. \ref{fig:system overview}, multi-modal sensor data are first captured within an adjustable time window, then aligned and aggregated through a unified data processing module (\ding{172}). Subsequently, the system extracts key features (\ding{173}) and embeds spatial, motion, and scene information into structured prompts (\ding{174}). Next, these prompts are fed into an LLM for context-aware activity inference (\ding{175}), producing two outputs: detailed activity logs (\ding{176}) and high-level activity summaries (\ding{177}).


\subsection{Data Collection and Processing}
{\systemname}extracts contextual semantic information of activities from various sensors; however, the data formats and sampling rates of these heterogeneous sensors vary significantly. To aggregate and align these data, we first systematically classify sensors. As shown in Table~\ref{tab:sensor}, these sensors are primarily categorized into four types based on their semantic meanings: spatial, activity, environmental, and physiological indicators. We then design a unified data processing model that formats data streams, obtains location information through reverse geocoding, semantically annotates environmental sensor data, and automatically updates motion and health readings.

\begin{table}[h]
    \centering
\resizebox{0.48\textwidth}{!}{\begin{tabular}{@{}ll@{}}
\toprule
\begin{tabular}[c]{@{}l@{}}\textbf{Spatial Context Awareness Sensors}\end{tabular}         & \begin{tabular}[c]{@{}l@{}}GPS, Barometer, Wi-Fi and Bluetooth.\end{tabular}                                                  \\ \midrule
\begin{tabular}[c]{@{}l@{}}\textbf{Motion State Monitoring Sensors}\end{tabular}           & \begin{tabular}[c]{@{}l@{}} Accelerometers, Gyroscopes and Magnetometers\end{tabular}                                   \\ \midrule
\begin{tabular}[c]{@{}l@{}}\textbf{Environmental Parameter Detection Sensors}\end{tabular} & \begin{tabular}[c]{@{}l@{}}Temperature, Microphone and Light sensors.\end{tabular}                                               \\ \midrule
\begin{tabular}[c]{@{}l@{}}\textbf{Physiological Indicator Monitoring Sensors}\end{tabular} & \begin{tabular}[c]{@{}l@{}}GSR, PPG, Thermistor and Infrared sensors.\end{tabular} \\ \bottomrule
\end{tabular}}
    \caption{Categorizing sensors on smartphones and watches. (GSR: Galvanic Skin Response, PPG: Photoplethysmography.)}
    \label{tab:sensor}
\end{table}

\subsubsection{Sensor Stream Formatting}
{\systemname}periodically reads and processes raw sensor data through a data processing module, utilizing an adjustable time window (e.g., 2 minutes). The processed data is subsequently converted into timestamped narratives and stored in a JSON format to facilitate efficient retrieval and readability. Sensor data collected from mobile devices are initially recorded using Unix timestamps, which represent the number of seconds that have elapsed since \textit{00:00:00 UTC} on \textit{January 1, 1970}. Based on the user's current time zone, {\systemname}then converts these timestamps into human-readable date formats.

\subsubsection{Reverse Geocoding and Localization}

To ensure accurate and reliable positioning under various environmental conditions, {\systemname}first extracts GPS coordinates and barometric readings, using the barometric formula \cite{wiki:barometric2025} to estimate the current altitude. It then employs the Google Maps API to convert GPS coordinates into structured address information, such as $[street, district, city, country]$, for more precise geographic context. Additionally, to compensate for GPS failures and reduce localization errors in low-signal environments (e.g., indoors), {\systemname}records wireless access point information detected within the same time window, including Wi-Fi SSIDs and Bluetooth MAC addresses. For frequently visited activity zones, such as home or office buildings, {\systemname}can refine spatial localization down to specific building floors or even rooms, if location information for deployed Wi-Fi access points or Bluetooth beacons is available.

\subsubsection{Semantic Annotation of Environmental Parameters}
To improve the interpretability of raw environmental data and enable more effective analysis and log generation, {\systemname}introduces a symbolic semantic annotation mechanism. By applying empirically defined thresholds, the system transforms the raw numerical data of light, audio, and temperature into semantically meaningful information, making it more interpretable and actionable. For example, the ambient light sensor, typically located near the screen of a smartphone, measures ambient light intensity (in Lux). Based on empirical knowledge, {\systemname}defines five semantic levels of illumination as follows: \textit{Level 1 (0–5 Lux)}: Extremely dark, such as a bedroom with the lights off or outdoor nighttime environments without artificial lighting; \textit{Level 2 (5–50 Lux)}: Dim, including settings with faint night lights in low-light indoor environments; \textit{Level 3 (50–300 Lux)}: Moderate brightness, commonly found in indoor daytime settings or softly lit office spaces; \textit{Level 4 (300–1000 Lux)}: Bright, such as near-window office areas or large commercial spaces; \textit{Level 5 ($>$1000 Lux)}: Harsh light, corresponding to outdoor environments under direct sunlight, such as midday on clear days.
For audio sensing, {\systemname}measures the decibel level relative to full scale (dBFS) to quantify the deviation of audio waveforms from the system’s maximum amplitude. This measurement provides an estimate of the ambient noise level. Using the dBFS metric, the system defines the following thresholds to categorize environmental sound levels:
$<$\textit{–70} \textit{dBFS}: Very Quiet. Near silent environments with almost no detectable sound.
\textit{(–70, –50) dBFS}: Soft Sound. There are some mild and gentle sounds, but the overall environment is quiet.
\textit{(–50, –30) dBFS}: Normal Sound. Typical ambient noise levels found in everyday life scenarios.
\textit{(–30, –10) dBFS}: Noisy. Environments with considerable noise, such as busy streets or crowded restaurants.
$>$ \textit{–10 dBFS}: Very Noisy. Highly noisy settings, including high traffic roads or concert venues.
In terms of temperature sensing, {\systemname}defines the following semantic temperature levels: $<$\textit{10}$^\circ\mathrm{C}$/\textit{50}$^{\circ}\mathrm{F}$: Cold; \textit{10–18$^\circ\mathrm{C}$}/\textit{50-64.4}$^{\circ}\mathrm{F}$: Cool; \textit{18–26$^\circ\mathrm{C}$}/\textit{64.4-78.8}$^{\circ}\mathrm{F}$: Comfortable; \textit{26–30$^\circ\mathrm{C}$}/\textit{78.8-86}$^{\circ}\mathrm{F}$: Warm; $>$\textit{30$^\circ\mathrm{C}$}/\textit{86}$^{\circ}\mathrm{F}$: Hot.  

\subsubsection{Refreshment of Motion and Physiological Sensing} 
{\systemname}acquires IMU motion data within specified time windows and extracts representative features, storing only the resulting feature vectors. This method meets the needs of activity recognition and analysis while minimizing memory usage by avoiding the storage of redundant raw data. The specific feature extraction strategy is described in detail in \textit{Sec.\ref{sec:LLM model}}. These features are crucial for accurately inferring users’ motion states and form the foundation for building complex, context-aware models. For physiological indicator sensors, most of which typically operate in a passive or periodically triggered mode, {\systemname}retains only the most recent data point within each time window for further analysis.

\subsection{Context-Aware Activity Understanding}\label{sec:LLM model}


LLMs excel at capturing complex patterns and relationships within unstructured data and generating rich and coherent expressions. Leveraging this advantage, we design an LLM-based context awareness module to produce high-quality semantic descriptions of activity contexts. 
We first propose a feature extraction mechanism that derives key features from raw activity sensor and audio data, transforming them into structured representations with rich semantic meaning. These interpretable representations not only compress lengthy time-series data effectively but also project the data into a textual space where LLMs excel, thereby enabling accurate context-aware activity reasoning. Building on this, we subsequently design a dynamic prompt generator. The generator embeds the extracted features into context-aware prompts, guiding the LLM to engage in structured reasoning and perform complex inference tasks. By carefully designing the prompt structure, contextual content, and task-specific constraints, our approach substantially improves both the reasoning accuracy and response consistency of {\systemnamenospace}.

\subsubsection{Feature Extraction}\label{sec:Feature Extraction} 
{\systemname}designs a lightweight feature extraction method to enable efficient and accurate activity context recognition. By integrating time-domain, frequency-domain, directional, and periodic features, the system can quickly extract key characteristics of human activities and scene changes during streaming processing. At runtime, {\systemname}first calculates the vector amplitude of the three-axis accelerometer, gyroscope, and magnetometer signals to comprehensively capture the overall strength and direction of motion. Based on this vector amplitude, it extracts four categories of key features in real time: \textbf{a) Nine time-domain statistical features}, including mean, standard deviation, skewness, kurtosis, maximum, minimum, quartiles, signal entropy, and temporal entropy, which characterize the temporal fluctuation and complexity of motion signals. \textbf{b) Six frequency-domain features}, comprising the logarithmic energy across five frequency bands and spectral entropy, reflecting the energy distribution patterns of different activities in the frequency domain.
\textbf{c) Two autocorrelation features}, designed to capture the periodic behavior of signals, thereby facilitating the differentiation of activities with repetitive motion patterns, such as walking and cycling.
\textbf{d) Nine axis-level statistical features derived from the raw triaxial signals}, including the mean and standard deviation of each axis, along with the inter-axis correlation coefficients. These features preserve directional information and enhance the accuracy of motion recognition.

This feature design spans both the temporal and frequency domains, while also capturing information on directionality and periodicity. The goal is to provide the LLM with a comprehensive perceptual representation, enhancing its ability to distinguish and interpret various types of human motion. Simultaneously, for audio data, {\systemname}extracts 120-dimensional feature vectors from the raw recordings. The first 60 dimensions include: 20 Mel-frequency cepstral coefficients (MFCCs) \cite{davis1980comparison} that capture the static spectral characteristics of ambient sounds, 20 Delta MFCCs that describe short-term dynamic changes, and 20 Acceleration MFCCs that reflect the rate of change in audio energy. The mean values of these three feature groups are utilized to model the spectral characteristics of the surrounding environment. The remaining 60 dimensions represent the standard deviations of the three feature groups, offering a measure of signal variability. This helps enhance the model's ability to detect changes in the environment.

\begin{figure}[t]
\centering
\begin{tcolorbox}[title=\textbf{Prompt: Activity Context Awareness Instruction}, 
  colback=gray!1, colframe=black!50, fonttitle=\bfseries, sharp corners=south]
  \begin{itemize}
  \item \textbf{Data Introduction:} You're an expert in signal analysis. Please analyze and predict the user's activity context based on the features of these sensor data... 
  \item \textbf{Feature Explanation:}
  \begin{itemize}
      \item Date-time:...  Location: ...
      \item Features meaning of motion and scene:
  \end{itemize}
  \item \textbf{Task Explanation:} Please give description of the activity context, including time, location, type and scene....
  \item \textbf{Specific Features vectors:} IMU:..., Audio:...
\item \textbf{Output Format:} Date-time:...; location information:...; activity category:...; scenario:...
\end{itemize}
\end{tcolorbox}
\caption{Structured prompt for Context Activity Awareness.}
\label{fig:structured-prompt}
\vspace{-0.2in}
\end{figure}

\begin{figure}[t]
\centering
\begin{tcolorbox}[title=\textbf{Prompt: Activity Log Summarization Instruction}, 
  colback=gray!1, colframe=black!50, fonttitle=\bfseries, sharp corners=south]

Please summarize the user's activity over the past few hours based on each activity log entry. Your summary should include:
a). The user's movement trajectory and locations visited.
    b). Changes in activity types and time distribution.
    c). Descriptions of environmental conditions (e.g., lighting, temperature, noise levels).
   d). Overall trends in physiological indicators.\\

Then, analyze the data to detect any of the following anomalies, and provide natural, conversational feedback to the user to raise awareness or suggest timely adjustments:
a). Environmental anomalies: prolonged exposure to extreme darkness, heat, or high noise levels.
b). Behavioral anomalies: extended periods of inactivity.
c). Health anomalies: elevated heart rate, low blood oxygen levels, or abnormal body temperature.
Please summarize and analyze step by step to produce a natural, logical piece of text.
\end{tcolorbox}
\caption{Structured prompt for Activity Log Summarization.}
\label{fig: summary_prompt}
\vspace{-0.2in}
\end{figure}

\subsubsection{Prompt Generation} 
Through extensive exploration of prompt engineering and iterative refinement, we design a structured prompt generator composed of five modular components: data introduction, feature explanation, task explanation, specific feature vectors, and output format,  as illustrated in Fig.~\ref{fig:structured-prompt}. Leveraging this structured prompt design, {\systemname}effectively guides the LLM to analyze sensor data patterns based on the semantic interpretations of various features, enabling accurate reasoning about the activity context. Moreover, {\systemname}incorporates location information extracted from GPS data and Wi-Fi/Bluetooth connectivity into the prompt generator. It guides the LLM to generate a detailed description of the activity location by leveraging its extensive general knowledge, including the specific address, an introduction to the place, and the location type, e.g., library, restaurant, hospital, residential area, computer science department. This rich contextual information plays a critical role in interpreting and summarizing users’ activity trajectories and daily behavioral patterns. Moreover, it substantially enhances the model’s ability to predict mobility patterns and activity scenarios based on location context. For example, when a user is at a shopping mall or subway station during the daytime, walking is more likely to occur, whereas being at home during nighttime increases the likelihood of lying down. Thus, {\systemname}aggregates and translates information from different sensor flow patterns into semantic space, guiding the LLM to make more accurate and comprehensive judgments about the active context combined with its intrinsic knowledge, which traditional classification models often struggle to achieve.

\subsection{Activity Log Generation and Summarization} 
To efficiently generate activity logs and summaries, {\systemname}applies Low-Rank Adaptation (LoRA) \cite{hu2022lora} to fine-tune an LLM model, enabling full deployment of the system on local personal computers and mobile devices. \subsubsection{Logs Summarization and Prompt Structure}{\systemname}integrates the activity contexts inferred by the LLM with previously generated semantic annotations from other sensor modalities to construct context-rich, structured activity logs.
Over a predefined extended time window (e.g., 2 hours, 4 hours, etc.), {\systemname}conducts semantic analysis and generates summaries for each fine-grained activity log entry. Leveraging the summarization capabilities of LLMs, {\systemname}conducts multi-level reasoning tasks to analyze the user's activity trajectory, main activities locations, surrounding environmental states, transitions in activity types, and physiological indicators. {\systemname}then focuses on abnormal patterns in the environment, behavior, and health signals, and timely sends a reminder to the user. For instance, if {\systemname}detects that the user has remained in environments characterized by high temperatures, low light, or excessive noise for an extended period, it infers potential risks and prompts the user to adjust their surroundings accordingly. Similarly, when abnormal physiological signals are detected (e.g., elevated heart rate or increased body temperature), {\systemname}generates personalized health warnings. The corresponding prompt template is shown in Fig. \ref{fig: summary_prompt}.


\subsubsection{Efficient Local Fine-tuning} 

To perform efficient reasoning while preserving user data privacy, {\systemname}explores local deployment strategies and fine-tunes pre-trained LLM model.
Among various open-source models, DeepSeek \cite{liu2024deepseek}, pretrained on large-scale and high-quality data, demonstrates superior performance in many context reasoning and text organization tasks. Balancing performance and resource constraints, we fine-tuned the DeepSeek R1-1.5B model \cite{liu2024deepseek}, whose smaller size facilitates deployment on resource-limited local devices.
{\systemname}first freezes the pretrained weights of the model and injects trainable low-rank adapters into the attention matrices, significantly reducing the number of tunable parameters. This approach ensures fast convergence while preserving most of the pretrained model's representational capacity. Furthermore, the system optimizes memory management by dynamically adjusting the batch size and optimizing memory caching. It automatically adapts its inference strategy based on available memory resources to prevent memory overflow issues. Finally, for deployment on the Raspberry Pi platform, {\systemname}compresses the model via 6-bit quantization to enable efficient inference. Based on these optimization strategies, {\systemname}enables fully local and efficient inference.



\section{Evaluation}
In this section, we first evaluate {\systemname}'s performance on three key tasks: location description, activity prediction, and scene understanding, using seven widely adopted benchmark datasets. Achieving good performance on these tasks is fundamental to the accurate generation of comprehensive activity logs. 
Subsequently, we construct a comprehensive daily activity dataset to evaluate {\systemnamenospace}'s performance on log generation and summarization tasks. 

\subsection{Experimental Setup}
\subsubsection{Datasets}



\textbf{Benchmark Datasets:} 
\textit{\textbf{a)}} \textit{StudentLife dataset} \cite{wang2014studentlife}: It comprises ten weeks of continuous activity location data collected from 48 students at Dartmouth College.
\textbf{\textit{b)}} \textit{HHAR} \cite{stisen2015smart}, \textit{Motion} \cite{malekzadeh2019mobile}, \textit{Shoaib} \cite{shoaib2014fusion} and \textit{UCI} \cite{reyes2016transition}: 
They are widely used datasets for human activity recognition, containing IMU sensor data gathered from various devices and participants.
\textit{\textbf{c)}} \textit{DCASE A} \cite{Mesaros2016_EUSIPCO} and \textit{DCASE B} \cite{DCASE2017challenge}: They are two acoustic scene classification datasets from the international DCASE Challenge, designed to support research and benchmarking in acoustic scene recognition tasks.
Table ~\ref{tab:Benchmark Datasets} presents the specific categories of activities/scenes and the number of classes included in the benchmark datasets.

\textbf{Comprehensive Activity Dataset Construction:}
Since no existing public dataset encompasses all four contextual dimensions: location, motion, environment, and physiology, we propose a set of composition rules and constraints to construct a comprehensive activity context dataset by integrating multiple established benchmarks \cite{wang2014studentlife,reyes2016transition,Mesaros2016_EUSIPCO,schmidt2018introducing}. 
To more accurately simulate real-world daily activity patterns, we design probabilistic rules conditioned on temporal and spatial factors, as well as dependencies between activities and scenes. 
Specifically, \textbf{\textit{a)}} During nighttime (00:00–08:00), activity probabilities are assigned as follows: 80\% lying, and 5\% each for sitting, standing, walking, and stair activities; \textbf{\textit{b)}} during daytime (08:01–23:59), the distribution is adjusted to 5\% lying, 50\% sitting, 20\% standing, 15\% walking, and 10\% stair activities. To simulate environmental transitions, when a location change is detected, audio data is randomly sampled from a different scene category.
\textbf{\textit{c)}} Dependencies between activities and scene transitions are defined as follows: Lying is followed by either sitting or standing; stair activities by walking; sitting transitions to standing or lying; and location changes between consecutive timestamps trigger walking.
\textbf{\textit{d)}} Finally, each activity sample was augmented with temporally trend aligned physiological signals from \cite{schmidt2018introducing}, including electrodermal activity (EDA), heart rate (HR), inter-beat interval (IBI), and temperature (TEMP). We will publicly release this dataset to support future research in this area.

\begin{table}[h]
\centering
\begin{minipage}[t]{0.47\textwidth}
\captionsetup{type=table}
\centering
\resizebox{1\textwidth}{!}{\begin{tabular}{@{}llc@{}}
\toprule
\textbf{Dataset} & \textbf{Activity or Scenario Categories}                                                                                                                                                                          & \multicolumn{1}{l}{\textbf{Classes}} \\ \midrule
\textbf{UCI} \cite{reyes2016transition}    & Standing, Sitting, Lying, Walking, Ascending Stairs, and Descending Stairs                                                                                                                               & 6                          \\
 \textbf{HHAR} \cite{stisen2015smart}   & Biking, Sitting, Standing, Walking, Ascending Stairs, and Descending Stairs                                                                                                                              & 6                          \\
\textbf{Motion} \cite{malekzadeh2019mobile} & Ascending Stairs, Descending Stairs, Walking, Jogging, Sitting, and Standing                                                                                                                             & 6                          \\
 \textbf{Shoaib} \cite{shoaib2014fusion} & Walking, Sitting, Standing, Jogging, Biking, Ascending Stairs, and Descending Stairs                                                                                                                     & 9                          \\ \midrule
\textbf{DCASE A\&B}  & Beach, Cafe/Restaurant, City Center, Forest Path, Metro Station, Tram, Park,        & \multirow{2}{*}{15} \\                        

 \cite{Mesaros2016_EUSIPCO,DCASE2017challenge} &  Residential Area, Home, Bus, Grocery Store, Car, Train, Office, Library        &    
\\ \bottomrule
\end{tabular}}
\vfill
\caption{Activity and Scenario Benchmark Datasets}
\label{tab:Benchmark Datasets}
\end{minipage}%
\end{table}

\begin{figure*}[t]
    \centering
    \includegraphics[width=\linewidth]{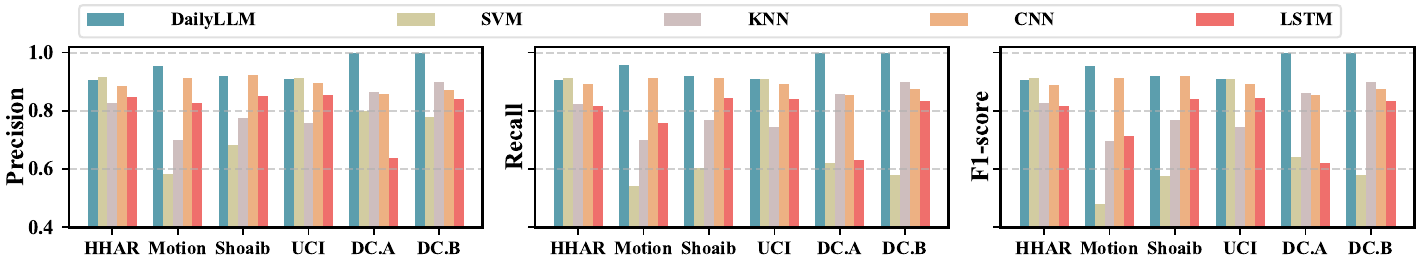}
    \caption{Performance comparison of {\systemname}with some machine learning (SVM \cite{tran2016humanSVM}, KNN \cite{mohsen2021human}) and deep learning (CNN \cite{munzner2017cnn}, LSTM \cite{chen2016lstm}) baseline methods on activity and scenario benchmark datasets (DC.A: DCASE A, DC.B: DCASE B).}
    \label{fig:overrall_performance}
\end{figure*}

\begin{table*}[t]
\centering
\resizebox{1\textwidth}{!}{\begin{tabular}{l|ccc|ccc|ccc|ccc|l|ccc|l|ccc}
\toprule
        \textbf{} & \multicolumn{3}{c|}{HHAR \cite{stisen2015smart}} 
        & \multicolumn{3}{c|}{Motion \cite{malekzadeh2019mobile}} 
        & \multicolumn{3}{c|}{Shoaib \cite{shoaib2014fusion}} 
        & \multicolumn{3}{c|}{UCI \cite{reyes2016transition}}  &  
        \multicolumn{1}{c|}{\textbf{}} & \multicolumn{3}{c|}{DCASE A \cite{Mesaros2016_EUSIPCO}} 
          & \multicolumn{1}{c|}{\textbf{}} & \multicolumn{3}{c}{DCASE B \cite{DCASE2017challenge}} \\
       \textbf{SOTA \textit{vs.} Ours} & \underline{Prec.}$(\uparrow)$ & \underline{Rec.} & \underline{F1} 
        & \underline{Prec.}$(\uparrow)$ & \underline{Rec.}$(\uparrow)$ & \underline{F1}$(\uparrow)$
        & \underline{Prec.}$(\uparrow)$ & \underline{Rec.}$(\uparrow)$ & \underline{F1}$(\uparrow)$
        & \underline{Prec.}$(\uparrow)$ & \underline{Rec.}$(\uparrow)$ & \underline{F1}$(\uparrow)$ & \textbf{SOTA \textit{vs.} Ours}
        & \underline{Prec.}$(\uparrow)$ & \underline{Rec.}$(\uparrow)$ & \underline{F1}$(\uparrow)$ &
        \textbf{SOTA \textit{vs.} Ours} & \underline{Prec.}$(\uparrow)$ & \underline{Rec.}$(\uparrow)$ & \underline{F1}$(\uparrow)$ \\
\midrule
LLaSA \cite{imran2024llasa}    & 0.88 & 0.86 & 0.84 & 0.85 & 0.83 & 0.83 & 0.84 & 0.81 & 0.81 & 0.82 & 0.75 & 0.72 & NMF \cite{BisotA} & 0.87 & 0.87 & 0.87 & ConvNet \cite{HanB} & 0.93 & 0.92 & 0.92 \\
\textbf{DailyLLM} & \textbf{0.91} &\textbf{ 0.90} & \textbf{0.91} & \textbf{0.96} & \textbf{0.96} & \textbf{0.96} & \textbf{0.92} & \textbf{0.92} & \textbf{0.92} & \textbf{0.91} & \textbf{0.91} & \textbf{0.91} &  \textbf{DailyLLM} & \textbf{1.00} & \textbf{1.00} &\textbf{ 1.00} & \textbf{DailyLLM} & \textbf{1.00} & \textbf{1.00} & \textbf{1.00} \\
\bottomrule
\end{tabular}}
\caption{Performance comparison of DailyLLM with SOTA results on activity and scenario benchmark datasets (Prec.: Precision, Rec.: Recall, F1: F1 Score). 
The SOTA results for the activity benchmark dataset come from LLaSA \cite{imran2024llasa}, while the SOTA results (NMF \cite{BisotA} and ConvNet \cite{HanB}) for the scenario dataset from the DCASE competition leaderboard \cite{Mesaros2016_EUSIPCO,DCASE2017challenge}.}
\label{tab:overall_performance}
\vspace{-0.2in}
\end{table*}

\subsubsection{Implementation}
{\systemname}is deployed and evaluated on a personal computer (NVIDIA RTX 4090 GPU, 24 GB) and a Raspberry Pi 5 \cite{raspberrypi5} (ARM Cortex-A76, 8 GB). For each benchmark dataset used in the activity and scene recognition tasks, we use only 500 and 200 labeled samples per class for training, respectively, while the remaining data is reserved for testing. For the location description task, we extracted 411 distinct locations from StudentLife for evaluation. The log generation and summarization task is evaluated on a comprehensive dataset, which contains a total of 8,896 samples, with a training to testing split ratio of 8:2.

\subsubsection{Evaluation Metrics}
\textbf{\textit{a)}} For activity prediction, scene understanding, and location description tasks, we adopt standard classification metrics: accuracy, precision, recall, and F1-score. 
\textbf{\textit{b)}} For log generation and summarization tasks, we employ BERTScore \cite{zhang2019bertscore} to quantify semantic similarity between generated outputs and ground-truth references, and use G-Eval \cite{liu2023g} to assess summary quality across five dimensions: accuracy, coverage, conciseness, consistency, and clarity. 

\subsubsection{Baselines}
\textbf{\textit{a)}} \textbf{Machine learning methods}: Support Vector Machine (SVM) \cite{tran2016humanSVM}, K-Nearest Neighbors (KNN) \cite{mohsen2021human}. \textbf{\textit{b)}} \textbf{Deep learning models}: Convolutional Neural Network (CNN) \cite{munzner2017cnn}, Long Short-Term Memory (LSTM) \cite{chen2016lstm}. We first compare the {\systemnamenospace}'s performance against classical machine learning and deep learning models on activity prediction and scene understanding tasks.
\textbf{\textit{c)}} \textbf{SOTA methods}: LLaSA (Large Language and Sensor Assistant) \cite{imran2024llasa} is a recent multi-modal large language model combining LIMU-BERT and LLaMA architectures, achieving leading performance on the four activity prediction benchmarks  \cite{stisen2015smart, malekzadeh2019mobile, shoaib2014fusion, reyes2016transition}.
Non-negative Matrix Factorization (NMF) \cite{BisotA} and ConvNet models \cite{HanB} represent the top-performing methods on the DCASE challenge datasets A \cite{Mesaros2016_EUSIPCO} and B \cite{DCASE2017challenge}, respectively, achieving the best results across 15 acoustic scene recognition tasks.
AutoLife \cite{xu2024autolife} is the SOTA method for log generation. However, as its code is not publicly released, we implement its inference framework and conduct a comparative evaluation against {\systemnamenospace}.

\subsection{Overall Performance}
\subsubsection{Activity Prediction}
Fig. \ref{fig:overrall_performance} presents the precision, recall, and F1-score results of {\systemname}compared with classical machine learning and deep learning methods across 6 benchmark datasets. {\systemname}achieves the highest average performance, demonstrating strong reasoning capabilities for activity-sensing data. We observe that the performance of traditional machine learning methods and learning methods exhibit significant fluctuations. For instance, SVM achieves strong performance on the HHAR dataset (Precision: 91.53\%) but performs poorly on the Motion dataset (Precision: 47.92\%); LSTM achieves an F1-score of 84.43\% on the UCI dataset but drops to 71.36\% on the Motion dataset. This highlights the sensitivity of these methods to variations in data. 
In contrast, {\systemname}leverages semantic context reasoning to deliver consistent and stable performance across datasets, showing better adaptability. Table~\ref{tab:overall_performance} compares {\systemname}with the recent LLM-based SOTA method LLaSA, where {\systemname}improves the average F1-score by 12.24\%. Moreover, {\systemname}achieves better performance than LLaSA, despite using a significantly smaller model with only 1.5B parameters compared to LLaSA’s 13B model, highlighting the effectiveness of our system design.

\begin{figure*}[t]
  \raggedright
  \raisebox{0in}{
  \begin{subfigure}[b]{0.48\textwidth}
    \centering
\resizebox{\textwidth}{!}{\begin{tabular}{@{}l|l|l|lll|c@{}}
\toprule
               & \multicolumn{1}{c|}{\begin{tabular}[c]{@{}c@{}}Activity\\  Prediction\end{tabular}} & \multicolumn{1}{c|}{\begin{tabular}[c]{@{}c@{}}Scene\\ Understanding\end{tabular}} & \multicolumn{3}{c|}{\begin{tabular}[c]{@{}c@{}}Logs Generation   \\ \underline{BERTScore ($\uparrow$)}\end{tabular}} & \multicolumn{1}{l}{Summary} \\ \midrule
LLM            & Acc. ($\uparrow$)                                                                            & Acc. ($\uparrow$)                                                                          & \multicolumn{1}{l|}{Prec. ($\uparrow$)}        & \multicolumn{1}{l|}{Rec. ($\uparrow$)}        & F1. ($\uparrow$)       & G-eval ($\uparrow$)                           \\ \midrule
Deepseek3 1.5B   & 81.50\%                                                                             & 91.30\%                                                                             & \multicolumn{1}{l|}{81.65\%}          & \multicolumn{1}{l|}{85.38\%}       & 83.47\%        & 0.8063                              \\ \midrule

Gemma3 4B    & 87.83\%                                                                             & 97.35\%                                                                            & \multicolumn{1}{l|}{96.56\%}          & \multicolumn{1}{l|}{95.52\%}         & 96.02\%        & 0.8115                              \\ \midrule
Qwen2.5 7B     & 86.83\%                                                                             & 97.26\%                                                                            & \multicolumn{1}{l|}{96.60\%}          & \multicolumn{1}{l|}{96.92\%}       & 96.76\%        & 0.8322                             \\ \midrule
Llama3 8B      & 88.67\%                                                                             & 99.32\%                                                                            & \multicolumn{1}{l|}{97.77\%}          & \multicolumn{1}{l|}{97.59\%}       & 97.86\%        & 0.8500                              \\ \midrule

GPT 4o    & 88.77\%                                                                             & 99.15\%                                                                            & \multicolumn{1}{l|}{98.54\%}          & \multicolumn{1}{l|}{98.43\%}       & 98.48\%        & 0.8840                              \\ 
 \bottomrule
\end{tabular}}
    \caption{}
    \label{tab:LLMs}
  \end{subfigure}}
 \begin{subfigure}[b]{0.24\textwidth}
    \centering
    \includegraphics[width=0.85\linewidth]{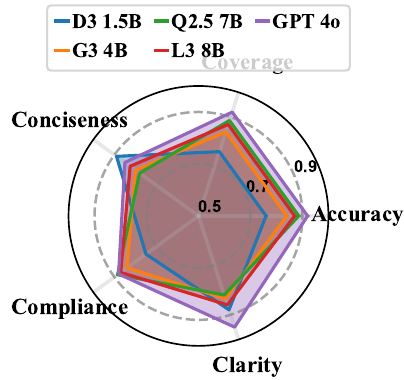}
    \caption{}
    \label{fig:RadarChart1}
  \end{subfigure}
  \begin{subfigure}[b]{0.24\textwidth}
    \centering
    \includegraphics[width=0.85\linewidth]{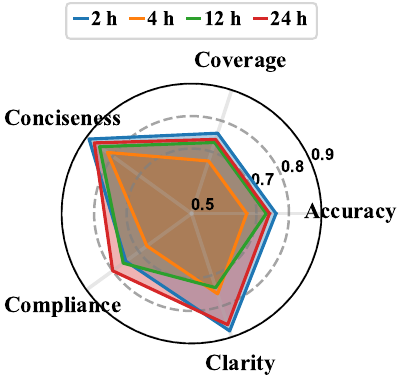}
    \caption{}
    \label{fig:RadarChart2}
  \end{subfigure}
  \caption{(a) System performance with different LLM Backbone. (b) Summary score with different LLM Backbone. (c) Summary score on different time window. (D3: Deepseek3, G3: Gemma3, Q2.5: Qwen2.5, L3: Llama3 8B.) }
  \vspace{-0.2in}
\end{figure*}

\subsubsection{Scene Understanding}
Fig.~\ref{fig:overrall_performance} and Table~\ref{tab:overall_performance} present the comparison of {\systemname}against baseline and SOTA methods for scene understanding on two acoustic scene datasets from DCASE \cite{Mesaros2016_EUSIPCO,DCASE2017challenge}.
The results demonstrate that {\systemname}achieves a 100\% accuracy rate in recognizing 15 distinct acoustic scenes, outperforming all baseline methods. 
Compared to SOTA methods (NMF and ConvNet), which are more prone to misclassifying similar scenes, {\systemname}guides the LLM to transfer general audio knowledge to the scene understanding task by focusing on key feature variations. With only 200 samples per class for fine-tuning, {\systemname}achieves superior performance, validating its effectiveness.



\subsubsection{Location Description}
We validate the accuracy of the location descriptions generated by {\systemname}using search engines and map information queries. Among 411 records, {\systemname} achieved an overall accuracy of 92.46\%.
This promising result indicates that {\systemname}has implicitly acquired substantial knowledge related to geography and infrastructure during its pre-training process, which is a task that general language models often struggle to perform effectively \cite{decoupes2024evaluation}.
To compare with SOTA LLM baselines, we input the raw GPS coordinates and Wi-Fi SSIDs into GPT-4o \cite{achiam2023gpt} for evaluation. The results show that GPT-4o often produced vague or incorrect outputs, exhibiting a high degree of hallucination. Only 72 descriptions are accurate, yielding an accuracy of 17.52\%, which highlights the necessity of {\systemnamenospace}'s design in generating context-aware location descriptions.


\subsubsection{Logs Generation and Summarization}
For activity log generation, {\systemname}achieves a BERTScore of Precision = 81.65\%, Recall = 85.38\%, and F1 = 83.47\%. 
Compared to SOTA baseline AutoLife's best performance (Precision = 65.0\%, Recall = 78.2\%, F1 = 70.4\%), {\systemname}demonstrates a 17\% improvement on precision.
Moreover, it integrates more sensor data to capture richer activity context, including environmental and physiological changes beyond AutoLife’s capabilities.
For summarization, we employ the general framework G-Eval~\cite{liu2023g} to assess the quality of the summaries. 
The results under the 2-hour window show that the overall quality score of the summary is 0.81. 
According to the G-Eval scores reported across various benchmarks~\cite{liu2023g}, the performance of {\systemname}'s summary can be considered Good.

\subsection{System Training and Inference Latency} We evaluate system's runtime on PC (NVIDIA RTX 4090, 24 GB) and Raspberry Pi 5 \cite{raspberrypi5} (ARM Cortex-A76, 8 GB). On the PC, training on 7,116 samples takes 5.24 hours. Inference on 1,780 samples averages 0.08s (activity prediction), 0.13s (scene understanding), and 2.22s (log generation). Summarization generation takes 1.63s for 2h time window.
Under the same setup, the SOTA baseline AutoLife generates activity logs in an average of 22.09 s. In comparison, {\systemname}achieves nearly a 10$\times$ speedup in inference. On the Raspberry Pi, the average reasoning times are 37 seconds for activity prediction, 90 seconds for scene understanding, 180 seconds for log generation, and 240 seconds for summarizing two-hour logs. These results suggest that even on resource-constrained mobile devices, {\systemname}is capable of generating high-quality activity summaries within four minutes, highlighting its significant potential for practical applications.

\subsection{Ablation Study}

\subsubsection{Feature Extraction}
To evaluate the impact of the feature extraction strategy on model performance, we conduct two ablation studies. First, we input raw IMU time-series data from the UCI dataset \cite{reyes2016transition} into {\systemname}for training, which results in a classification accuracy of only 20.8\%. It is substantially lower than the 91\% accuracy achieved using our approach.
We also train {\systemname}for acoustic scene understanding using only 13 basic MFCC features extracted from DCASE dataset A \cite{Mesaros2016_EUSIPCO}. This setup achieves an average accuracy of 64.2\%. 



\subsubsection{Prompt Engineering}
To evaluate the impact of prompt design, we compared our carefully crafted prompt with two alternative configurations: \textit{a) Naive Prompt}, which only provides the most essential feature vectors and task description; and \textit{b) Free-form Prompt}, which expresses the task and data freely in natural language without explicit constraints or guidance for step-by-step reasoning or structured prompts. Results indicate a significant performance drop for {\systemname}under both settings. When using the Naive Prompt, the accuracy for activity prediction and scene understanding dropped from 91\% and 100\% to 35.4\% and 56.8\%, respectively. Under the Free-form Prompt, the accuracies decreased to 67.4\% and 75.3\% for the two tasks, respectively.

\subsection{Sensitivity Analysis}
\subsubsection{Time Windows}
We evaluate the log summarization performance of {\systemname}under different time window settings (2, 4, 12, and 24 hours). As shown in Fig. \ref{fig:RadarChart2}, the summarization scores remain relatively stable across these four configurations, without significant degradation. 
However, due to the context length limitations of current LLMs, it is not feasible to include an unlimited number of activity logs. As the time window increases, the system must adopt a coarser downsampling strategy for data selection. Therefore, to balance information integrity and processing efficiency, a 2-hour time window is set as the system’s default.



\subsubsection{LLM Backbones}
To assess {\systemname}performance with different LLM backbones, we test five mainstream models with varying  parameter sizes and calculate the average results.
As shown in Fig~\ref{tab:LLMs}, {\systemname}performs well with all models. 
Based on GPT-4o, {\systemname}can achieve the best overall results, with log generation F1-score of 98.48\%. As illustrated in Fig.~\ref{fig:RadarChart1}, it also achieves the highest score 0.88 in the summarization task. However, GPT-4o is not open-source, requires data to be uploaded to external servers, and incurs a cost for reasoning over each token. With Gemma-3 4B as backbones, {\systemname}also demonstrates strong performance, but its hard to deployment on mobile devices.
Therefore, to balance cost, efficiency, and performance, {\systemname}adopts the lightweight DeepSeek-R1-1.5B model as its LLM backbone.


\section{Conclusion}

Activity log generation is an important research problem, yet existing approaches still face limitations. In this paper, we present {\systemnamenospace}, a novel system driven by LLMs and multi-modal sensors, provides an effective solution for automatically recording and summarizing human activities.
Compared to SOTA methods, {\systemname}achieves a 17\% improvement in precision while delivering a 10$\times$ increase in inference efficiency. Even on resource-constrained edge devices (e.g., Raspberry Pi 5), it can summarize 2 hours activity data within 4 minutes.
Extensive experiments demonstrate the strong potential and practical effectiveness of {\systemname} in real-world.


\section*{acknowledgements}
This work has been funded in part by NSF, with award numbers \#2112665, \#2112167, \#2003279, \#2120019, \#2211386, \#2052809, \#1911095 and in part by PRISM and CoCoSys, centers in JUMP 2.0, an SRC program sponsored by DARPA.


\bibliographystyle{IEEEtran}
\bibliography{references}

\begin{thebibliography}{10}
\providecommand{\url}[1]{#1}
\csname url@samestyle\endcsname
\providecommand{\newblock}{\relax}
\providecommand{\bibinfo}[2]{#2}
\providecommand{\BIBentrySTDinterwordspacing}{\spaceskip=0pt\relax}
\providecommand{\BIBentryALTinterwordstretchfactor}{4}
\providecommand{\BIBentryALTinterwordspacing}{\spaceskip=\fontdimen2\font plus
\BIBentryALTinterwordstretchfactor\fontdimen3\font minus \fontdimen4\font\relax}
\providecommand{\BIBforeignlanguage}[2]{{%
\expandafter\ifx\csname l@#1\endcsname\relax
\typeout{** WARNING: IEEEtran.bst: No hyphenation pattern has been}%
\typeout{** loaded for the language `#1'. Using the pattern for}%
\typeout{** the default language instead.}%
\else
\language=\csname l@#1\endcsname
\fi
#2}}
\providecommand{\BIBdecl}{\relax}
\BIBdecl

\bibitem{statistaSmartphone}
\BIBentryALTinterwordspacing
{Statista}. (2025) Global smartphone penetration 2016-2023. [Online]. Available: \url{https://www.statista.com/statistics/203734/global-smartphone-penetration-per-capita-since-2005/}
\BIBentrySTDinterwordspacing

\bibitem{aiordachioae2019life}
A.~Aiordachioae and R.-D. Vatavu, ``Life-tags: a smartglasses-based system for recording and abstracting life with tag clouds,'' \emph{Proceedings of the ACM on human-computer interaction}, vol.~3, pp. 1--22, 2019.

\bibitem{jiang2019memento}
S.~Jiang, Z.~Li, P.~Zhou, and M.~Li, ``Memento: An emotion-driven lifelogging system with wearables,'' \emph{ACM Transactions on Sensor Networks (TOSN)}, vol.~15, no.~1, pp. 1--23, 2019.

\bibitem{aiorduachioae2024integrating}
A.~Aiord{\u{a}}chioae, A.~Calinciuc, and M.~D. Schipor, ``Integrating extended reality and neural headsets for enhanced emotional lifelogging: A technical overview,'' in \emph{2024 International Conference on Development and Application Systems (DAS)}.\hskip 1em plus 0.5em minus 0.4em\relax IEEE, 2024, pp. 1--6.

\bibitem{xu2024autolife}
H.~Xu, P.~Tong, M.~Li, and M.~Srivastava, ``Autolife: Automatic life journaling with smartphones and llms,'' \emph{arXiv:2412.15714}, 2024.

\bibitem{post2025contextllm}
K.~Post, R.~Kuchida, M.~Olapade, Z.~Yin, P.~Nurmi, and H.~Flores, ``Contextllm: Meaningful context reasoning from multi-sensor and multi-device data using llms,'' in \emph{Proceedings of ACM HOTMOBILE'25}.\hskip 1em plus 0.5em minus 0.4em\relax Association for Computing Machinery (ACM), 2025.

\bibitem{lifelogapp}
\BIBentryALTinterwordspacing
G.~Asano, ``Lifelog: Timelog \& diary,'' 2024, 2024. [Online]. Available: \url{https://apps.apple.com/bb/app/lifelog-timelog-diary/id6473384260}
\BIBentrySTDinterwordspacing

\bibitem{dayOneApp}
\BIBentryALTinterwordspacing
{Automattic}. (2025) Day one journal app | your journal for life. [Online]. Available: \url{https://dayoneapp.com/}
\BIBentrySTDinterwordspacing

\bibitem{sahoo2019foodai}
D.~Sahoo, W.~Hao, S.~Ke, W.~Xiongwei, H.~Le, P.~Achananuparp, E.-P. Lim, and S.~C. Hoi, ``Foodai: Food image recognition via deep learning for smart food logging,'' in \emph{Proceedings of the 25th ACM SIGKDD International Conference on Knowledge Discovery \& Data Mining}, 2019, pp. 2260--2268.

\bibitem{jeong2020cyberslacking}
Y.~Jeong, H.~Jung, and J.~Lee, ``Cyberslacking or smart work: Smartphone usage log-analysis focused on app-switching behavior in work and leisure conditions,'' \emph{International Journal of Human--Computer Interaction}, vol.~36, no.~1, pp. 15--30, 2020.

\bibitem{krieter2018analyzing}
P.~Krieter and A.~Breiter, ``Analyzing mobile application usage: generating log files from mobile screen recordings,'' in \emph{Proceedings of the 20th international conference on human-computer interaction with mobile devices and services}, 2018, pp. 1--10.

\bibitem{lewin2024novel}
D.~Lewin, C.~M. Starling, E.~S. Zhou, D.~Greenberg, C.~Shaw, and H.~Arem, ``A novel voice interactive sleep log: concurrent validity with actigraphy and sleep diaries,'' \emph{Journal of Clinical Sleep Medicine}, vol.~20, no.~2, pp. 309--312, 2024.

\bibitem{hu2022lora}
E.~J. Hu, Y.~Shen, P.~Wallis, Z.~Allen-Zhu, Y.~Li, S.~Wang, L.~Wang, W.~Chen \emph{et~al.}, ``Lora: Low-rank adaptation of large language models.'' \emph{ICLR}, vol.~1, no.~2, p.~3, 2022.

\bibitem{imran2024llasa}
S.~A. Imran \emph{et~al.}, ``Llasa: Large multimodal agent for human activity analysis through wearable sensors,'' \emph{arXiv preprint arXiv:2406.14498}, 2024.

\bibitem{kok2024iot}
{\.I}.~K{\"o}k, O.~Demirci, and S.~{\"O}zdemir, ``When iot meet llms: Applications and challenges,'' in \emph{2024 IEEE International Conference on Big Data (BigData)}.\hskip 1em plus 0.5em minus 0.4em\relax IEEE, 2024, pp. 7075--7084.

\bibitem{an2024iot}
T.~An, Y.~Zhou, H.~Zou, and J.~Yang, ``Iot-llm: Enhancing real-world iot task reasoning with large language models,'' \emph{arXiv preprint arXiv:2410.02429}, 2024.

\bibitem{mo2024iot}
S.~Mo, R.~Salakhutdinov, L.-P. Morency, and P.~P. Liang, ``Iot-lm: Large multisensory language models for the internet of things,'' \emph{arXiv preprint arXiv:2407.09801}, 2024.

\bibitem{xu2024penetrative}
H.~Xu, L.~Han, Q.~Yang, M.~Li, and M.~Srivastava, ``Penetrative ai: Making llms comprehend the physical world,'' in \emph{Proceedings of the 25th International Workshop on Mobile Computing Systems and Applications}, 2024, pp. 1--7.

\bibitem{ji2024hargpt}
S.~Ji, X.~Zheng, and C.~Wu, ``Hargpt: Are llms zero-shot human activity recognizers?'' \emph{arXiv preprint arXiv:2403.02727}, 2024.

\bibitem{VirtualAnnotators}
\BIBentryALTinterwordspacing
A.~Hota, S.~Chatterjee, and S.~Chakraborty, ``Evaluating large language models as virtual annotators for time-series physical sensing data,'' \emph{ACM Trans. Intell. Syst. Technol.}, Sep. 2024. [Online]. Available: \url{https://doi.org/10.1145/3696461}
\BIBentrySTDinterwordspacing

\bibitem{berenguer2024using}
A.~Berenguer, A.~Morej{\'o}n, D.~Tom{\'a}s, and J.-N. Maz{\'o}n, ``Using large language models to enhance the reusability of sensor data,'' \emph{Sensors}, vol.~24, no.~2, p. 347, 2024.

\bibitem{wiki:barometric2025}
\BIBentryALTinterwordspacing
{Wikipedia contributors}, ``Barometric formula --- wikipedia{,} the free encyclopedia,'' 2025. [Online]. Available: \url{https://en.wikipedia.org/wiki/Barometric_formula}
\BIBentrySTDinterwordspacing

\bibitem{davis1980comparison}
S.~Davis and P.~Mermelstein, ``Comparison of parametric representations for monosyllabic word recognition in continuously spoken sentences,'' \emph{IEEE transactions on acoustics, speech, and signal processing}, vol.~28, no.~4, pp. 357--366, 1980.

\bibitem{liu2024deepseek}
A.~Liu, B.~Feng, B.~Xue, B.~Wang, B.~Wu, C.~Lu, C.~Zhao, C.~Deng, C.~Zhang, C.~Ruan \emph{et~al.}, ``Deepseek-v3 technical report,'' \emph{arXiv preprint arXiv:2412.19437}, 2024.

\bibitem{wang2014studentlife}
R.~Wang, F.~Chen, Z.~Chen, T.~Li, G.~Harari, S.~Tignor, X.~Zhou, D.~Ben-Zeev, and A.~T. Campbell, ``Studentlife: assessing mental health, academic performance and behavioral trends of college students using smartphones,'' in \emph{Proceedings of the 2014 ACM international joint conference on pervasive and ubiquitous computing}, 2014, pp. 3--14.

\bibitem{stisen2015smart}
A.~Stisen, H.~Blunck, S.~Bhattacharya, T.~S. Prentow, M.~B. Kj{\ae}rgaard, A.~Dey, T.~Sonne, and M.~M. Jensen, ``Smart devices are different: Assessing and mitigatingmobile sensing heterogeneities for activity recognition,'' in \emph{Proceedings of the 13th ACM conference on embedded networked sensor systems}, 2015, pp. 127--140.

\bibitem{malekzadeh2019mobile}
M.~Malekzadeh, R.~G. Clegg, A.~Cavallaro, and H.~Haddadi, ``Mobile sensor data anonymization,'' in \emph{Proceedings of the international conference on internet of things design and implementation}, 2019, pp. 49--58.

\bibitem{shoaib2014fusion}
M.~Shoaib, S.~Bosch, O.~D. Incel, H.~Scholten, and P.~J. Havinga, ``Fusion of smartphone motion sensors for physical activity recognition,'' \emph{Sensors}, vol.~14, no.~6, pp. 10\,146--10\,176, 2014.

\bibitem{reyes2016transition}
J.-L. Reyes-Ortiz, L.~Oneto, A.~Sam{\`a}, X.~Parra, and D.~Anguita, ``Transition-aware human activity recognition using smartphones,'' \emph{Neurocomputing}, vol. 171, pp. 754--767, 2016.

\bibitem{Mesaros2016_EUSIPCO}
A.~Mesaros, T.~Heittola, and T.~Virtanen, in \emph{24th European Signal Processing Conference 2016 (EUSIPCO 2016)}, Budapest, Hungary, 2016.

\bibitem{DCASE2017challenge}
A.~Mesaros, T.~Heittola, A.~Diment, B.~Elizalde, A.~Shah, E.~Vincent, B.~Raj, and T.~Virtanen, ``{DCASE} 2017 challenge setup: Tasks, datasets and baseline system,'' in \emph{Proceedings of the Detection and Classification of Acoustic Scenes and Events 2017 Workshop (DCASE2017)}, November 2017, pp. 85--92.

\bibitem{schmidt2018introducing}
P.~Schmidt, A.~Reiss, R.~Duerichen, C.~Marberger, and K.~Van~Laerhoven, ``Introducing wesad, a multimodal dataset for wearable stress and affect detection,'' in \emph{Proceedings of the 20th ACM international conference on multimodal interaction}, 2018, pp. 400--408.

\bibitem{tran2016humanSVM}
D.~N. Tran and D.~D. Phan, ``Human activities recognition in android smartphone using support vector machine,'' in \emph{2016 7th international conference on intelligent systems, modelling and simulation (isms)}.\hskip 1em plus 0.5em minus 0.4em\relax IEEE, 2016, pp. 64--68.

\bibitem{mohsen2021human}
S.~Mohsen, A.~Elkaseer, and S.~G. Scholz, ``Human activity recognition using k-nearest neighbor machine learning algorithm,'' in \emph{Proceedings of the International Conference on Sustainable Design and Manufacturing}.\hskip 1em plus 0.5em minus 0.4em\relax Springer, 2021, pp. 304--313.

\bibitem{munzner2017cnn}
S.~M{\"u}nzner, P.~Schmidt, A.~Reiss, M.~Hanselmann, R.~Stiefelhagen, and R.~D{\"u}richen, ``Cnn-based sensor fusion techniques for multimodal human activity recognition,'' in \emph{Proceedings of the 2017 ACM international symposium on wearable computers}, 2017, pp. 158--165.

\bibitem{chen2016lstm}
Y.~Chen, K.~Zhong, J.~Zhang, Q.~Sun, and X.~Zhao, ``Lstm networks for mobile human activity recognition,'' in \emph{2016 International conference on artificial intelligence: technologies and applications}.\hskip 1em plus 0.5em minus 0.4em\relax Atlantis Press, 2016, pp. 50--53.

\bibitem{BisotA}
V.~Bisot, R.~Serizel, S.~Essid, and G.~Richard, ``Supervised nonnegative matrix factorization for acoustic scene classification,'' DCASE2016 Challenge, Tech. Rep., September 2016.

\bibitem{HanB}
Y.~Han and J.~Park, ``Convolutional neural networks with binaural representations and background subtraction for acoustic scene classification,'' DCASE2017 Challenge, Tech. Rep., September 2017.

\bibitem{raspberrypi5}
\BIBentryALTinterwordspacing
{Raspberry}, ``Raspberry pi 5,'' 2025. [Online]. Available: \url{https://www.raspberrypi.com/products/raspberry-pi-5/}
\BIBentrySTDinterwordspacing

\bibitem{zhang2019bertscore}
T.~Zhang, V.~Kishore, F.~Wu, K.~Q. Weinberger, and Y.~Artzi, ``Bertscore: Evaluating text generation with bert,'' \emph{arXiv preprint arXiv:1904.09675}, 2019.

\bibitem{liu2023g}
Y.~Liu, D.~Iter, Y.~Xu, S.~Wang, R.~Xu, and C.~Zhu, ``G-eval: Nlg evaluation using gpt-4 with better human alignment,'' \emph{arXiv preprint arXiv:2303.16634}, 2023.

\bibitem{decoupes2024evaluation}
R.~Decoupes, R.~Interdonato, M.~Roche, M.~Teisseire, and S.~Valentin, ``Evaluation of geographical distortions in language models,'' in \emph{International Conference on Discovery Science}.\hskip 1em plus 0.5em minus 0.4em\relax Springer, 2024, pp. 86--100.

\bibitem{achiam2023gpt}
J.~Achiam, S.~Adler, S.~Agarwal, L.~Ahmad, I.~Akkaya, F.~L. Aleman, D.~Almeida, J.~Altenschmidt, S.~Altman, S.~Anadkat \emph{et~al.}, ``Gpt-4 technical report,'' \emph{arXiv preprint arXiv:2303.08774}, 2023.

\end{thebibliography}



\end{document}